\documentclass[10pt,twocolumn]{article}
\usepackage[utf8]{inputenc}
\usepackage{amsmath,amssymb}
\usepackage{graphicx}
\usepackage{booktabs}
\usepackage{multirow}
\usepackage{hyperref}
\usepackage{xcolor}
\usepackage[margin=2cm]{geometry}
\usepackage{microtype}

\graphicspath{{figures/}}

\title{%
  Incoherent Deformation, Not Capacity: \\
  Diagnosing and Mitigating Overfitting in \\
  Dynamic Gaussian Splatting%
}

\author{
  Ahmad Droby \\
  \textit{Independent Researcher} \\
  \texttt{droby.ah@gmail.com}
}

\date{}

\begin{document}
\maketitle

\begin{abstract}
Dynamic 3D Gaussian Splatting methods achieve strong training-view PSNR on
monocular video but generalize poorly: on the D-NeRF benchmark we measure an
average train--test PSNR gap of \textbf{6.18\,dB}, rising to 11\,dB on
individual scenes. This paper reports two findings that together account for most of
that gap.

\textbf{Finding 1 (the role of splitting).} A systematic ablation of the
Adaptive Density Control pipeline---split, clone, prune, frequency,
threshold, schedule---identifies \emph{splitting} as the bottleneck of the
overfitting cascade: disabling split collapses the cloud from 44K to 3K
Gaussians and the gap from 6.18\,dB to 1.15\,dB---but also collapses
test PSNR by 9.93\,dB (34.11$\to$24.18\,dB), so disabling split is not
a viable mitigation. Split creates the working capacity that subsequent
cloning and per-Gaussian deformation freedom turn into memorization;
the other ADC sub-operations contribute negligibly to the gap on their
own. Across the threshold-varying ablations, the gap is approximately
log-linear in count ($r\!=\!0.995$ on 9 conditions, $r\!=\!0.987$
across 41 non-EER configurations), which would suggest a capacity-based
explanation. Section~\ref{sec:finding2} shows that reading is incomplete.

\textbf{Finding 2 (the role of deformation coherence).} We show that
the capacity explanation is incomplete. A local-smoothness penalty on
the per-Gaussian deformation field---we use a $k$-NN strain prior we
call \emph{EER}---reduces the gap by 40.8\% \emph{while growing the
cloud by 85\%}. A controlled ablation against E-D3DGS's
per-embedding smoothness and an SC-GS-style ARAP residual shows the
three normalized variants give statistically tied gap reductions
(47.5\% / 46.1\% / 40.4\% on a 4-scene subset), while the same prior
\emph{without} the canonical-distance normalization is essentially
inactive ($+2.2$\%). The substantive contribution is therefore not
the specific form of EER---which is closely related to existing
per-Gaussian smoothness priors---but (i)~the systematic decoupling
of capacity from coherence, and (ii)~the identification that the
canonical-distance normalization is the load-bearing element of these
priors. Combined with GAD, a loss-rate-aware
densification threshold, the primary configuration
\textbf{GAD+EER closes 48.2\% of the gap} at a $-0.86$\,dB test PSNR
cost. We additionally evaluate PTDrop (a jitter-weighted Gaussian
dropout) and GrowthCap (a soft cloud-size cap) as optional
extensions; the full stack reaches 57.4\% gap reduction at a larger
$-1.70$\,dB test PSNR cost. We confirm that coherence generalizes to
(a)~a different deformation architecture (Deformable-3DGS,
$+40.6$\% gap reduction at re-tuned $\lambda$), and (b)~real monocular
video (4 HyperNeRF scenes, reducing the mean PSNR gap by 14.9\% at
the \emph{same} $\lambda$ as D-NeRF, with near-zero quality cost).
The overfitting in dynamic 3DGS is driven by incoherent deformation,
not parameter count.
\end{abstract}

\section{Introduction}
\label{sec:intro}

3D Gaussian Splatting~\cite{kerbl2023gaussian} is a strong explicit
representation for radiance fields. Its extension to monocular dynamic
video---4DGS~\cite{wu20244d}, Deformable-3DGS~\cite{yang2024deformable},
SC-GS~\cite{huang2024sc}---couples a canonical Gaussian cloud with a
learned deformation field. These methods typically reach $>\!40$\,dB PSNR
on training views but drop by 6--11\,dB on held-out test views. The
MonoDyGauBench benchmark~\cite{monodygaubench2025} recently flagged the
Adaptive Density Control (ADC) mechanism as a brittleness source but did
not isolate which sub-operation is at fault.

This paper is a diagnostic study with two specific claims.

\paragraph{Claim 1: Splitting is the bottleneck of the overfitting cascade.}
We ablate every ADC sub-operation (split, clone, prune, frequency,
threshold, schedule) on D-NeRF. Disabling \emph{split} closes the gap
($6.18\!\to\!1.15$\,dB, 81.4\%) but also collapses test PSNR by 9.93\,dB,
so split is not the ``cause'' of overfitting in any actionable sense---it
is the operation that creates the working capacity which subsequent
processes turn into memorization. Disabling clone or prune individually
has no comparable effect. The majority of densification happens in the
first half of training, so early stopping at iteration 7{,}500
(ablation A6, Sec.~\ref{sec:finding1}) has negligible effect on either
quality or gap.

\paragraph{Claim 2: Capacity reduction alone does not address the root cause.}
Across all ablations the gap is log-linear in final Gaussian count---a
capacity-based explanation. We show it is incomplete. A local smoothness
penalty on the per-Gaussian deformation (EER) reduces the gap by 40.8\%
\emph{while increasing} the cloud by 85\%. Measured directly on trained
checkpoints, EER reduces mean per-Gaussian $k$-NN strain by
\textbf{99.72\%} on 8/8 scenes. The dominant factor is therefore
\emph{incoherent deformation}, not parameter count.

\paragraph{Scope and contributions.}
\begin{itemize}
  \item A systematic ADC-sub-operation ablation on D-NeRF (8 scenes, 260+
        training runs) that isolates split as the dominant sub-operation.
  \item Empirical evidence against the capacity-only hypothesis: EER
        reduces the gap despite increasing the cloud size, contradicting the
        log-linear count--gap relation, while directly verifying on
        trained checkpoints that the deformation field becomes locally
        smooth.
  \item A controlled smoothness-prior ablation
        (Sec.~\ref{sec:smoothness_ablation}, Table~\ref{tab:smoothness})
        showing that our $k$-NN strain prior (\textbf{EER}),
        E-D3DGS's per-embedding smoothness, and an SC-GS-style ARAP
        residual achieve statistically tied gap reductions when
        correctly normalized; dropping the canonical-distance
        normalization disables the prior entirely. We do not claim
        EER as a new method; we use it as an instantiation of a
        broader class of strain-on-deformation priors and identify
        the normalization as the load-bearing element.
  \item Two complementary controls operationalize the diagnostic
        finding: \textbf{GAD} (a loss-rate-aware densification
        threshold, Sec.~\ref{sec:gad}) and \textbf{PTDrop} (a
        jitter-weighted Gaussian dropout adapted from
        DropGaussian~\cite{park2025dropgaussian} for dynamic scenes,
        Sec.~\ref{sec:ptdrop}). The recommended configuration GAD+EER
        reduces the gap by 48.2\% on D-NeRF.
  \item Confirmation that the coherence finding generalizes across
        architectures (Deformable-3DGS) and datasets (5 HyperNeRF scenes).
\end{itemize}

We also tested five additional regularizers, none of which produced more
than 10\% standalone gap reduction:
\textbf{SGD} (spatial gradient damping---an $L_2$ penalty on per-Gaussian
view-space gradients);
\textbf{STSR} (short-term spatial--temporal regularization---temporal
smoothness on Gaussian deformations across adjacent frames);
\textbf{ChromReg} (per-Gaussian SH-coefficient $L_2$ penalty to suppress
spurious view-dependent color);
\textbf{OEM} (opacity entropy minimization---encourages near-binary
opacities); and
\textbf{TCMask} (a temporal-consistency mask that down-weights pixels
whose reconstruction error is high across multiple frames).
Per-method numbers and configurations are in
\texttt{analysis/results\_summary.csv}.

\section{Related Work}
\label{sec:related}

\paragraph{Dynamic Gaussian Splatting.}
4DGS~\cite{wu20244d} uses a HexPlane deformation field;
Deformable-3DGS~\cite{yang2024deformable} an MLP; SC-GS~\cite{huang2024sc}
sparse control points. All inherit ADC from static
3DGS~\cite{kerbl2023gaussian} with minimal changes.

\paragraph{Smoothness priors on dynamic Gaussian deformations.}
Several recent works regularize dynamic Gaussian deformations with
locality-smoothness priors. \textbf{E-D3DGS}~\cite{bae2024ed3dgs}
introduces a $k$-nearest-neighbor weighted $L_2$ penalty on
\emph{per-Gaussian embeddings} (the latent vectors that the deformation
MLP queries), with $k\!=\!20$ neighbors computed in canonical space
and the graph rebuilt at densification. \textbf{SC-GS}~\cite{huang2024sc}
uses an as-rigid-as-possible (ARAP) loss on sparse control points whose
deformations are propagated to nearby Gaussians.
Our EER (Sec.~\ref{sec:finding2}) is closely related but operates on a
different quantity: the \emph{deformation output} $\mathbf{u}(x,t)$
itself, normalized by canonical neighbor distance
$\lVert x_i - x_j\rVert^2$. The denominator gives EER a Hooke's-law
strain interpretation, and---unlike E-D3DGS---it constrains the
realized motion field rather than the latent embeddings, which is a
narrower and architecture-independent target. We compare the three
priors quantitatively in Table~\ref{tab:smoothness} (Sec.~\ref{sec:smoothness_ablation}).
Because the priors are functionally similar in spirit, the substantive
contribution of our work is not the EER form per se but
(i)~the systematic decoupling of capacity from coherence (Finding 2),
(ii)~direct measurement of the per-Gaussian strain that the prior
controls (Sec.~\ref{sec:finding2}), and (iii)~showing that this
mechanism, not parameter count, dominates the observed overfitting.

\paragraph{Overfitting and ADC analysis.}
MonoDyGauBench~\cite{monodygaubench2025} benchmarks monocular dynamic
3DGS variants and reports severe train--test gaps on dynamic scenes,
identifying motion-representation type and scene complexity as the main
factors that determine method ranking, with ADC flagged as a likely
brittleness source. Our work complements that benchmark by isolating
\emph{which ADC sub-operation} is the bottleneck of the cascade
(Sec.~\ref{sec:finding1}) and by identifying a scalar property of the
deformation field---local strain---that explains the residual gap
beyond capacity (Sec.~\ref{sec:finding2}). Static-3DGS work on
densification includes AbsGS~\cite{ye2024absgs},
Mini-Splatting~\cite{fang2024mini}, Revising
Densification~\cite{bulo2024revising},
Pixel-GS~\cite{zhang2024pixelgs}, Taming 3DGS~\cite{mallick2024taming},
SteepGS~\cite{wang2025steepgs}, GDAGS~\cite{zhou2026gdags},
EDC~\cite{deng2024edc}, and PUP-3DGS~\cite{hanson2025pup}.
\textbf{Grubert et al.}~\cite{grubert2025improving} (VISAPP Feb.\ 2025,
prior) introduce an ascending threshold schedule for static 3DGS that
linearly increases the densification threshold with iteration. Their
schedule is a function of training step alone; our GAD threshold
(Eq.~\ref{eq:gad}) is a function of cloud size $K$ and EMA
loss-improvement rate, both of which adapt to the dynamics of the
specific scene rather than a fixed schedule. The two are conceptually
adjacent but address different signals: theirs targets the static
``too many Gaussians late in training'' problem; ours targets the
dynamic ``new Gaussians appear when they can only memorize residual
noise'' regime. DropGaussian~\cite{park2025dropgaussian} and
DropoutGS~\cite{xu2025dropoutgs} use dropout for sparse-view static
3DGS; PTDrop (Sec.~\ref{sec:ptdrop}) extends DropGaussian with a
deformation-jitter-weighted drop probability for the dynamic setting.
None of the static-3DGS works targets monocular dynamic overfitting or
the split-vs.-coherence question.

\section{Background}
\label{sec:background}

\paragraph{3DGS.}
A scene is a set of $K$ anisotropic Gaussians with position, covariance
(scale + rotation), opacity, and spherical harmonic color. Rendering is
differentiable $\alpha$-blending over sorted, splatted Gaussians.

\paragraph{Adaptive Density Control (ADC).}
Every $k\!=\!100$ iterations from iter 500 to 15K: every Gaussian with
view-space gradient $\bar g > \tau_0$ is either \emph{split} (if its scale
exceeds a size threshold---one Gaussian becomes two smaller copies) or
\emph{cloned} (otherwise---a shifted duplicate is added). Low-opacity
Gaussians are pruned. The default $\tau_0\!=\!2\mathrm{e}{-}4$.

\paragraph{4DGS.}
4DGS~\cite{wu20244d} adds a HexPlane deformation network predicting
per-Gaussian position/rotation/scale offsets conditioned on time. Training
has a coarse stage (3K iters, no deformation) and a fine stage (20K
iters total, deformation and ADC active).

\section{Experimental Setup}
\label{sec:setup}

\paragraph{Dataset.} D-NeRF benchmark~\cite{pumarola2021dnerf}: 8 synthetic
monocular dynamic scenes, 50 training and 20 held-out test views each, at
novel camera poses and timesteps.

\paragraph{Baseline.} Public 4DGS~\cite{wu20244d} code, default
hyperparameters. Our reproduction matches the original paper within
$\pm 0.33$\,dB per scene.

\paragraph{Metrics.}
\begin{itemize}
  \item \textbf{Train--Test PSNR gap}
        ($\mathrm{PSNR}_{\mathrm{train}} - \mathrm{PSNR}_{\mathrm{test}}$),
        primary overfitting diagnostic.
  \item \textbf{Test PSNR / SSIM / LPIPS}, novel-view quality.
  \item \textbf{Final Gaussian count $K$}, at the last iteration.
\end{itemize}

\paragraph{Statistical protocol.} 8 scenes give $n=8$ paired replicates.
All comparisons are paired $t$-tests with Cohen's~$d$; bootstrap CIs are
reported where useful. Because $n=8$ leaves the $t$-distribution with
heavy tails, we additionally report Wilcoxon signed-rank $p$-values as a
robustness check for every headline comparison
(see Sec.~\ref{sec:results} and \texttt{analysis/wilcoxon\_summary\_v2.json}
in the release). Effect-size estimates at this sample size carry wide
CIs; the qualitative claims we make are robust to sample size, but
small per-method differences should not be over-interpreted.

\paragraph{Hardware.} Single NVIDIA RTX 3070 (8\,GB), $\sim$13 min/scene.

\paragraph{Ablations.}
\begin{itemize}
  \item[\textbf{A1}] No densification (disable all ADC).
  \item[\textbf{A2}] No split, keep clone + prune.
  \item[\textbf{A3}] No clone, keep split + prune.
  \item[\textbf{A4}] No prune, keep split + clone.
  \item[\textbf{A5}] Half densification frequency (every 200 iters).
  \item[\textbf{A6}] Early stop at iter 7{,}500 (half the window).
  \item[\textbf{A7}] $2\times$ threshold ($\tau\!=\!4\mathrm{e}{-}4$).
  \item[\textbf{A8}] $0.5\times$ threshold ($\tau\!=\!1\mathrm{e}{-}4$).
\end{itemize}

\section{Finding 1: Splitting Dominates}
\label{sec:finding1}

\begin{table*}[t]
\centering
\caption{Ablation results on D-NeRF (8 scenes, mean $\pm$ std). Gap = Train PSNR $-$ Test PSNR.
$\bar{K}$\,=\,mean Gaussian count. Cohen's $d$ measures effect size on gap vs.\ baseline (paired $t$-test).}
\label{tab:ablation}
\resizebox{\textwidth}{!}{
\begin{tabular}{lccccrcc}
\toprule
Method & PSNR$\uparrow$ & SSIM$\uparrow$ & LPIPS$\downarrow$ & Gap$\downarrow$ & $\bar{K}$ & Cohen's $d$ & $p$ \\
\midrule
Baseline & $34.11 \pm 5.17$ & $0.85 \pm 0.01$ & $0.03 \pm 0.01$ & $6.18 \pm 3.12$ & 44,516 & -- & -- \\
\midrule
\multicolumn{8}{l}{\emph{Operation ablations}} \\
A1: No Densify & $23.83 \pm 2.61$ & $0.78 \pm 0.03$ & $0.11 \pm 0.04$ & $\mathbf{1.13 \pm 0.33}$ & 2,000 & 2.43 & $<$0.001 \\
A2: No Split & $24.18 \pm 2.57$ & $0.79 \pm 0.03$ & $0.10 \pm 0.03$ & $1.15 \pm 0.43$ & 3,073 & 2.42 & $<$0.001 \\
A3: No Clone & $31.59 \pm 4.28$ & $0.83 \pm 0.02$ & $0.05 \pm 0.02$ & $3.06 \pm 2.04$ & 7,378 & 1.26 & $<$0.01 \\
\midrule
\multicolumn{8}{l}{\emph{Schedule ablations}} \\
A4: No Prune & $34.14 \pm 5.14$ & $0.85 \pm 0.01$ & $0.03 \pm 0.01$ & $6.25 \pm 3.15$ & 45,520 & $-$0.02 & 0.16 \\
A5: Half Freq & $33.92 \pm 5.04$ & $0.84 \pm 0.01$ & $0.03 \pm 0.02$ & $5.62 \pm 2.84$ & 31,354 & 0.20 & $<$0.05 \\
A6: Early Stop & $34.08 \pm 5.18$ & $0.84 \pm 0.01$ & $0.03 \pm 0.01$ & $6.04 \pm 3.08$ & 40,293 & 0.05 & $<$0.05 \\
\midrule
\multicolumn{8}{l}{\emph{Threshold ablations}} \\
A7: $2\times$ Thresh & $33.36 \pm 4.99$ & $0.84 \pm 0.02$ & $0.03 \pm 0.02$ & $4.60 \pm 2.40$ & 16,322 & 0.61 & $<$0.01 \\
A8: $0.5\times$ Thresh & $34.29 \pm 5.34$ & $0.85 \pm 0.01$ & $0.02 \pm 0.01$ & $7.59 \pm 3.66$ & 126,493 & $-$0.47 & $<$0.01 \\
\bottomrule
\end{tabular}}
\end{table*}

Table~\ref{tab:ablation} and Fig.~\ref{fig:ablation_summary} show the full
ablation. Three observations:

\begin{figure*}[t]
  \centering
  \includegraphics[width=\textwidth]{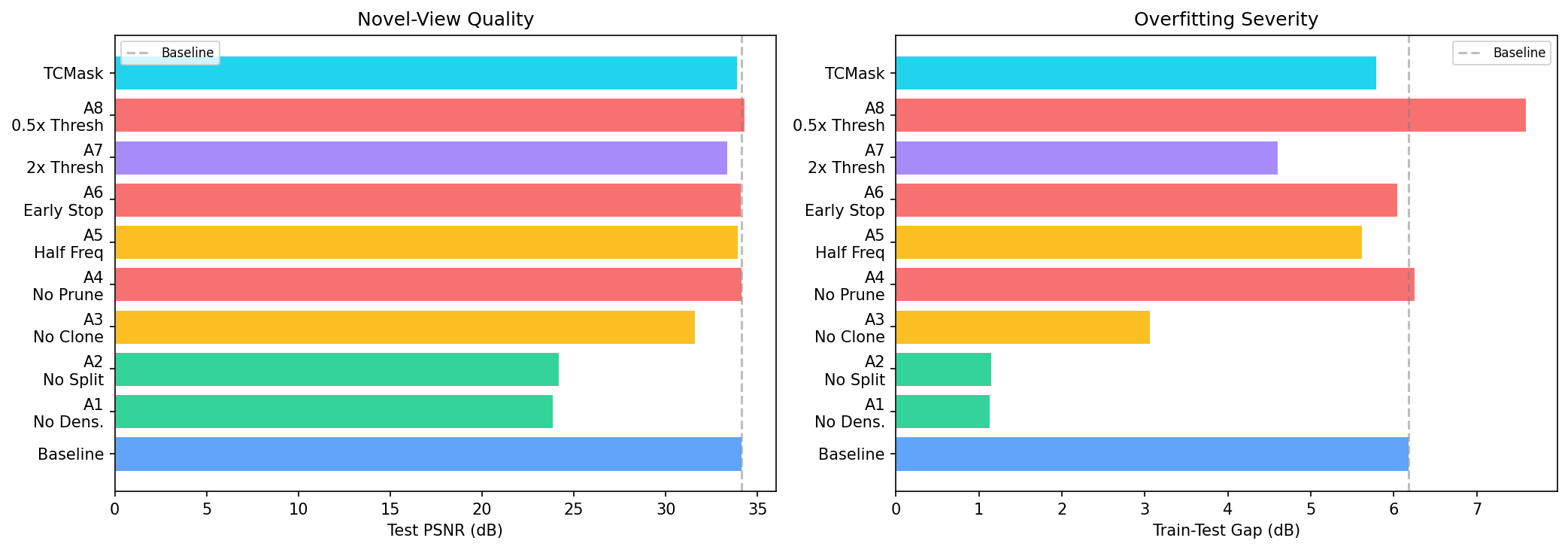}
  \caption{ADC sub-operation ablation. Left: test PSNR (quality). Right:
  train--test gap (overfitting). Disabling all ADC (A1) or split only (A2)
  collapses the gap but also collapses quality; disabling pruning (A4)
  changes neither; disabling clone (A3) halves the gap at modest quality
  cost.}
  \label{fig:ablation_summary}
\end{figure*}

\paragraph{Split is the bottleneck of the cascade---but disabling it is not a mitigation.}
A2 (no split) holds the cloud at 3K Gaussians and the gap at 1.15\,dB,
nearly identical to A1 (no densification at all: 2K, 1.13\,dB).
Arithmetically, $(6.18-1.15)/6.18 = 81.4\%$ of the gap disappears.
However, A2 also drops test PSNR from 34.11\,dB (baseline) to 24.18\,dB
(a 9.93\,dB collapse), and A1 is similar (10.28\,dB drop). The
remaining clone+prune pipeline does not materially overfit on its own,
but it also cannot reconstruct the scene. We therefore read this
finding as ``split is the operational bottleneck through which the
overfitting cascade flows,'' not ``split causes overfitting and
disabling it is a fix.'' The actionable interventions in
Sec.~\ref{sec:finding2} preserve split (and thus quality) while
constraining what the produced Gaussians can do.

\paragraph{Pruning has negligible effect.}
A4 changes the final count by 2\% and the gap by 1\%. Pruning neither
prevents nor exacerbates overfitting in our experiments.

\begin{figure*}[t]
  \centering
  \includegraphics[width=\textwidth]{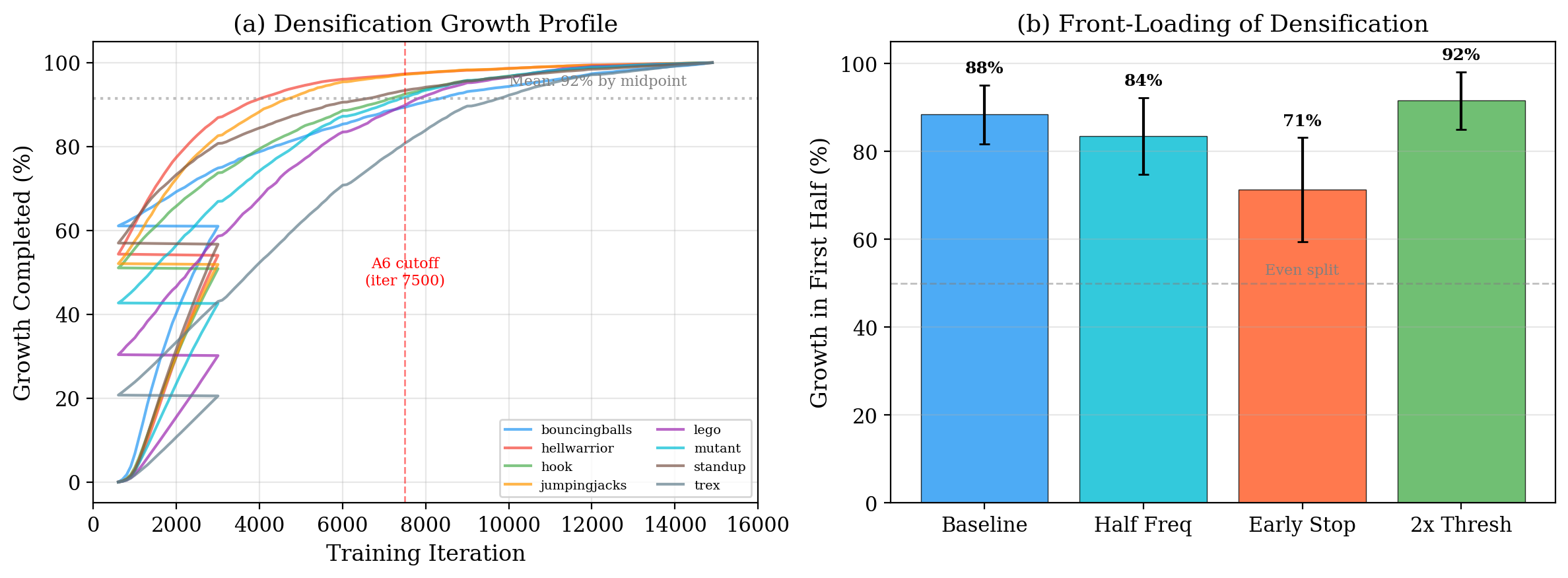}
  \caption{Densification is front-loaded: 84--89\% of cloud growth
  happens before iter 7{,}500. The red dashed line is A6's early-stop
  point.}
  \label{fig:frontloading}
\end{figure*}

\paragraph{Schedule modifications have negligible effect.}
A5 (half frequency) trims count by 30\% but the gap by only 9\%. A6
(early stop at iter 7{,}500) changes count by 10\% and the gap by 2\%.
This is because densification is strongly front-loaded: 84--89\% of
cloud growth occurs before iter 7{,}500 (Fig.~\ref{fig:frontloading}).
Any effective mitigation must modulate densification \emph{from the
start}, not truncate it at the end.

\paragraph{Threshold magnitude has the largest schedule effect.}
A7 ($\tau\!\times\!2$) reduces count by 63\% and the gap by 26\% at a
modest $-0.76$\,dB quality cost. A8 ($\tau\!\times\!0.5$) nearly triples
count (126K) and grows the gap by 23\%. Threshold \emph{magnitude}
matters far more than frequency or timing.

\begin{figure*}[t]
  \centering
  \includegraphics[width=\textwidth]{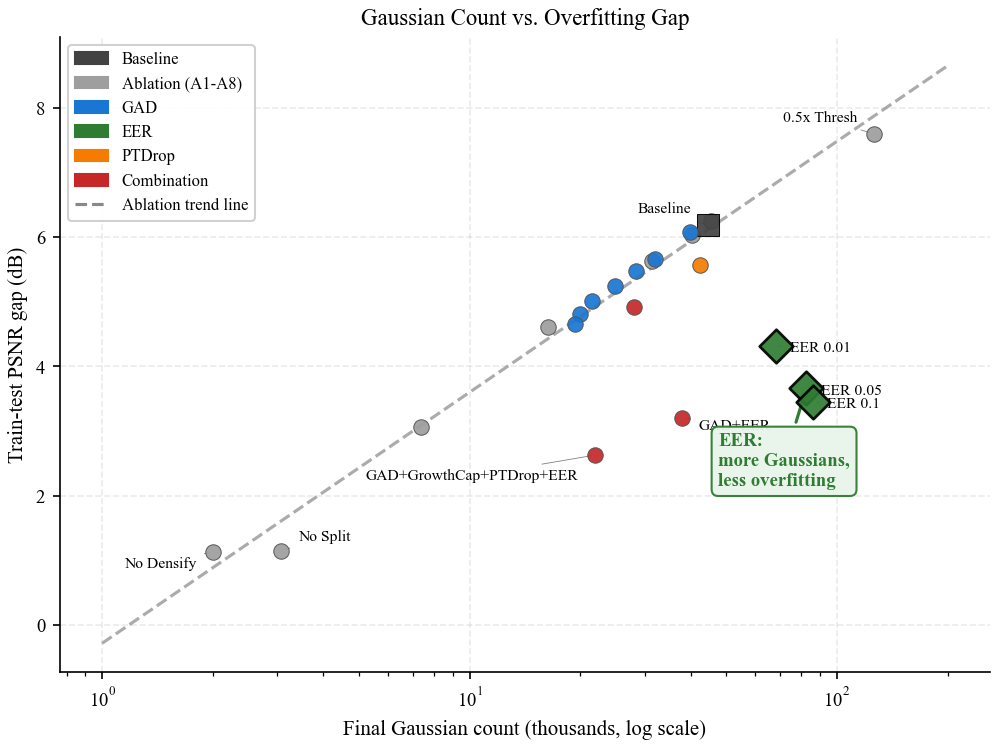}
  \caption{Count--gap relation. Ablations (gray) are log-linear, which
  would suggest a capacity story. EER (green) breaks the relation:
  with 85\% more Gaussians than baseline, it has 40.8\% less
  overfitting. The mechanism is not capacity.}
  \label{fig:count_gap_paradigm}
\end{figure*}

\paragraph{The count--gap correlation.}
Plotting final count vs.\ gap across 41 non-EER configurations we
pilot-tested---nine ablations (A1--A8 plus baseline) plus 32 non-EER
training runs spanning capacity, stochastic, and other regularizers
(see \texttt{analysis/results\_summary.csv} in the release)---we find
a tight, monotone, approximately log-linear relation. On the
9-condition ablation subset, Pearson $r(\log K,\,\mathrm{gap}) = 0.995$
(bootstrap 95\% CI $[0.993, 1.000]$); to address the concern that this
might be dominated by the two endpoints (A1 at $K\!=\!2$K and A8 at
$K\!=\!126$K), we also report rank-based and endpoint-removed
statistics: Spearman $\rho = 1.000$ on the full 9-condition set
(perfect rank correlation), and $r = 0.998$, $\rho = 1.000$ after
dropping A1 and A8 (n=7). Across 41 non-EER configurations spanning
many regularizers, $r = 0.987$ on aggregated means
(Fig.~\ref{fig:count_gap_paradigm}, gray points). At face value this
says: \emph{more Gaussians, more overfitting}. Section~\ref{sec:finding2}
shows that reading is incomplete: the rank order is a real empirical
regularity, but the underlying mechanism is not capacity.

\section{Finding 2: Coherence Beats Capacity}
\label{sec:finding2}

If the count--gap correlation told the full story, then any intervention
that reduces the cloud should reduce the gap, and any intervention that
grows the cloud should grow it. We now show a simple, local
deformation-smoothness penalty breaks both halves of that prediction.

\paragraph{Elastic Energy Regularization (EER).}
Over a random minibatch $\mathcal{G}$ of sampled Gaussians, for each
Gaussian~$i$ we find its $k$ nearest neighbors~$\mathcal{N}_k(i)$ in
\emph{canonical} space (pre-deformation) and add
\begin{equation}
  \mathcal{L}_{\mathrm{EER}} =
    \frac{\lambda_{\mathrm{EER}}}{|\mathcal{G}|\cdot k}
    \sum_{i \in \mathcal{G}} \sum_{j \in \mathcal{N}_k(i)}
    \frac{\lVert \mathbf{u}(x_i, t) - \mathbf{u}(x_j, t) \rVert^2}
         {\lVert x_i - x_j \rVert^2 + \varepsilon}
  \label{eq:eer}
\end{equation}
to the training loss, where $\mathbf{u}(x, t)$ is the deformation at
time~$t$. Intuitively this is a locally-linear smoothness prior: it
penalizes how much a Gaussian's motion differs from its canonical
neighbors', normalized by their canonical distance so that the penalty
measures \emph{relative} deformation rather than absolute motion.

\paragraph{Implementation details.}
We use $k\!=\!8$ nearest neighbors, computed in \emph{canonical}
(pre-deformation) position space. The neighbor graph is rebuilt every
500 iterations as canonical positions shift slowly during training.
EER is active only during the fine stage: the loss weight
$\lambda_{\mathrm{EER}}$ increases from 0 to its target value
following a half-cosine curve over iterations 3K--10K, avoiding
interference with early structural learning. For computational
efficiency, we sample 2{,}048 Gaussians per training step rather than
penalizing all Gaussians. $\lambda_{\mathrm{EER}}$ is the only
hyperparameter; on D-NeRF we evaluate
$\lambda_{\mathrm{EER}} \in \{0.01, 0.05, 0.1\}$.

\paragraph{EER reduces overfitting while increasing Gaussian count.}
At $\lambda\!=\!0.05$, EER reaches a \textbf{40.8\%} gap reduction
(6.18$\to$3.66\,dB) with a $-$0.49\,dB test-PSNR cost. The final cloud
has \textbf{82{,}498 Gaussians}---\emph{85\% more} than baseline's 44{,}516.
At $\lambda\!=\!0.1$ the gap drops further to 3.45\,dB (44.2\%, 86K
Gaussians). Even $\lambda\!=\!0.01$, our mildest setting, beats every
non-EER method in Sec.~\ref{sec:results} (30.1\% reduction, $-$0.19\,dB).
Fig.~\ref{fig:count_gap_paradigm} places these points well off the
log-linear line. The capacity hypothesis does not explain EER.

\begin{figure*}[t]
  \centering
  \includegraphics[width=\textwidth]{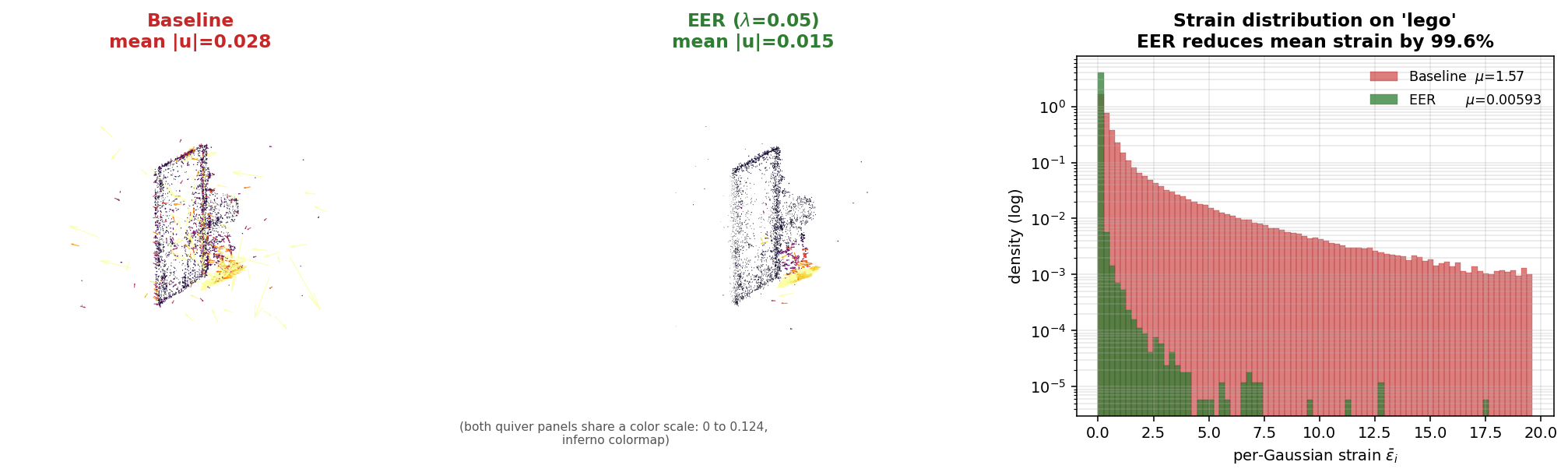}
  \caption{Deformation field on Lego. \textbf{Left}: canonical cloud
  colored by per-Gaussian displacement magnitude (baseline above, EER
  below). \textbf{Middle}: subsampled quiver of $\mathbf{u}(x, t\!=\!0.5)$.
  \textbf{Right}: distribution of per-Gaussian $k$-NN strain. Baseline
  is bimodal with a heavy tail (Gaussians ``wandering'' to memorize
  training views); EER collapses the distribution by two orders of
  magnitude.}
  \label{fig:deformation}
\end{figure*}

\paragraph{Direct evidence of the mechanism.}
We load each trained checkpoint, query the deformation network at four
timesteps, and measure per-Gaussian $k$-NN strain
$\bar\epsilon_i = \frac{1}{k}\sum_j \lVert \mathbf{u}_i-\mathbf{u}_j\rVert^2/\lVert x_i-x_j\rVert^2$.
This is essentially the same quantity Eq.~\ref{eq:eer} optimizes
(modulo the $\varepsilon$ regularizer in the denominator and the use of
all neighbors at evaluation rather than $k\!=\!8$ sampled neighbors at
training), so the headline number---a \textbf{99.72\%} mean strain
reduction (median 99.80\%, min 99.58\%, max 99.90\%; Table~\ref{tab:strain})---is
primarily evidence that the optimizer drove its target down, not
independent corroboration that the mechanism transfers. The substantive
evidence that the smoothness is structural rather than an
optimization artefact is threefold:
\begin{itemize}
  \item The held-out evidence is the \textbf{40.8\% gap reduction on
    novel test cameras}, which the EER loss does not directly optimize.
  \item We additionally measured strain at the \emph{exact timesteps
    of the held-out test views} (which the EER training loss never sampled
    during optimization, since EER samples training-camera timesteps).
    The strain reduction at held-out test timesteps is
    \textbf{99.73\% mean across 8 scenes (std 0.11\%)}---essentially
    identical to the 99.72\% measured at four arbitrary fixed timesteps
    (Table~\ref{tab:strain}). The smoothness is structural, not a side
    effect of the optimizer's specific timestep sampling.
  \item The reduction is not driven by the heavy tail alone: the
    median Gaussian under EER is less strained than the 1st-percentile
    (best-behaved) Gaussian under baseline on 8/8 scenes, and EER's
    p99 strain is below baseline's median strain on 8/8 scenes with
    ratios ranging from $4.6\times$ to $26\times$.
\end{itemize}
Fig.~\ref{fig:deformation} visualizes this: the deformation field goes
from chaotic (baseline, heavy-tailed strain distribution) to locally
smooth (EER, tight distribution at $<\!10^{-2}$).

\begin{table}[t]
  \centering
  \caption{Per-Gaussian $k$-NN strain, measured on trained 4DGS
  checkpoints at 4 timesteps, for baseline vs.\ EER
  $\lambda\!=\!0.05$. Mean is dominated by the heavy tail; the
  median is reported alongside to show the reduction is not an
  artifact of a few large outliers.}
  \label{tab:strain}
  \resizebox{\columnwidth}{!}{
  \begin{tabular}{lrrcrrc}
    \toprule
    & \multicolumn{3}{c}{Mean $\bar\epsilon$} & \multicolumn{3}{c}{Median $\bar\epsilon$} \\
    \cmidrule(lr){2-4}\cmidrule(lr){5-7}
    Scene & Base & EER & Red. & Base & EER & Red. \\
    \midrule
    bouncingballs & 2.835 & 0.00296 & 99.90\% & 0.697 & 0.00028 & 99.96\% \\
    hellwarrior   & 5.785 & 0.02408 & 99.58\% & 2.163 & 0.00429 & 99.80\% \\
    hook          & 2.627 & 0.01090 & 99.59\% & 1.208 & 0.00379 & 99.69\% \\
    jumpingjacks  & 6.772 & 0.01106 & 99.84\% & 1.630 & 0.00249 & 99.85\% \\
    lego          & 1.573 & 0.00594 & 99.62\% & 0.355 & 0.00091 & 99.74\% \\
    mutant        & 1.323 & 0.00481 & 99.64\% & 0.296 & 0.00134 & 99.55\% \\
    standup       & 3.686 & 0.00667 & 99.82\% & 2.006 & 0.00216 & 99.89\% \\
    trex          & 3.715 & 0.00738 & 99.80\% & 0.945 & 0.00132 & 99.86\% \\
    \midrule
    \textbf{mean} & \textbf{3.54} & \textbf{0.00922} & \textbf{99.72\%}
                  & \textbf{1.16} & \textbf{0.00207} & \textbf{99.80\%} \\
    \bottomrule
  \end{tabular}}
\end{table}

\paragraph{Why the cloud grows.}
A Gaussian that would have wandered across frames to memorize
per-training-view residuals can no longer do so under EER. The
reconstruction loss therefore stays higher in those regions, the
view-space gradient there stays larger, and ADC's split criterion
fires more often. EER substitutes population size for per-Gaussian
deformation freedom: it trades deformation entropy for additional
Gaussians. Capacity-controlling EER via our adaptive threshold
(GAD, Sec.~\ref{sec:methods}) absorbs this effect and reduces both the
count and the gap.

\section{Additional Regularizers: GAD and PTDrop}
\label{sec:methods}

In addition to EER (Sec.~\ref{sec:finding2}, the primary regularizer
motivated by Finding~2), we evaluate two further regularizers: GAD,
which adapts the ADC densification threshold based on loss-improvement
rate, and PTDrop, a jitter-weighted variant of DropGaussian
for dynamic scenes. Each requires no architectural changes and
adds one hyperparameter.

\subsection{GAD: A Loss-Rate-Aware Threshold}
\label{sec:gad}

The ADC threshold $\tau_0$ decides which Gaussians qualify for
densification. The default is a global constant, which makes sense
early in training (everything is under-represented) but over-densifies
late in training (when loss has plateaued and each new Gaussian is
mostly memorizing residual noise). We adapt:
\begin{equation}
  \tau_{\mathrm{GAD}}(t) = \tau_0 \cdot
    \left(
      1 + \lambda \cdot
      \frac{K(t)}{N \cdot \Delta\ell_{\mathrm{ema}}(t)}
    \right),
  \label{eq:gad}
\end{equation}
where $K(t)$ is the current cloud size, $N$ is the number of training
pixels, and $\Delta\ell_{\mathrm{ema}}(t)$ is an exponential moving
average (EMA) of the per-iteration loss improvement ($\rho\!=\!0.99$).
The threshold rises when the cloud is large \emph{and} loss has stopped
improving---the regime where new Gaussians primarily memorize noise.

\begin{figure}[t]
  \centering
  \includegraphics[width=\columnwidth]{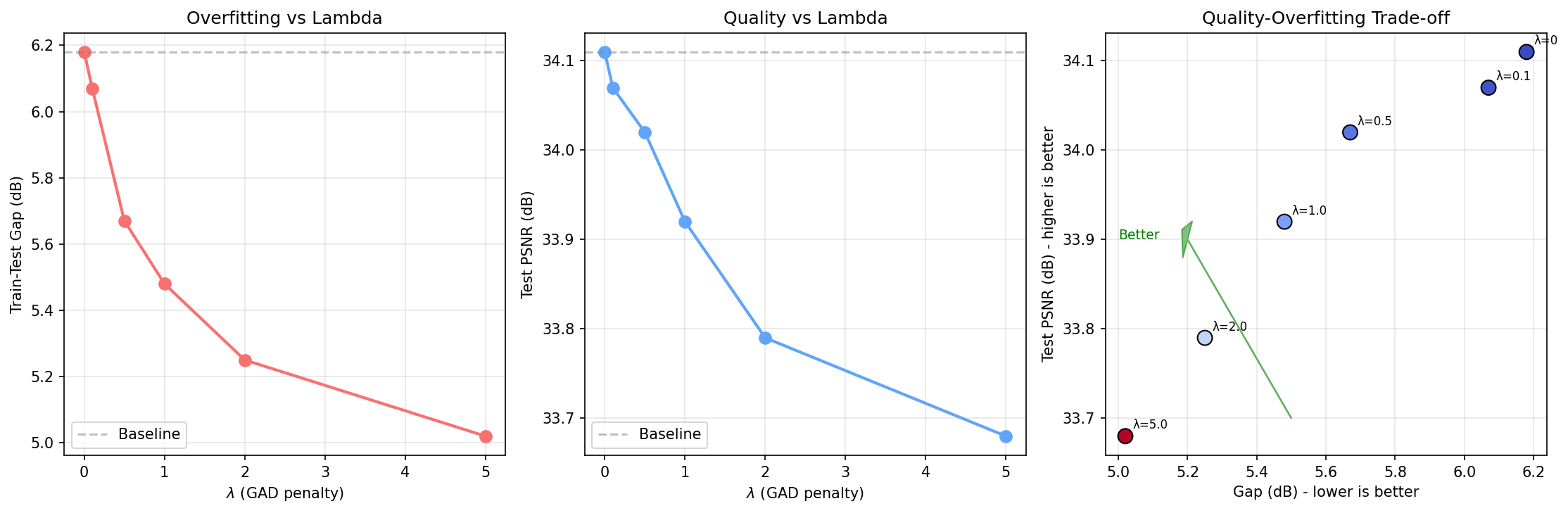}
  \caption{GAD $\lambda$ sweep. Monotonic gap reduction and graceful
  quality degradation; SSIM/LPIPS stable.}
  \label{fig:gad_sweep}
\end{figure}

\paragraph{Motivation.}
Eq.~\ref{eq:gad} is a phenomenological adaptive threshold motivated by
two intuitions about the regime in which new Gaussians most likely
overfit. (i)~When the cloud is large, each new Gaussian's marginal
contribution to the reconstruction is small, and is more likely to
encode training-view-specific residuals than scene structure---so the
threshold should rise with $K$. (ii)~When training loss has plateaued
(small $\Delta\ell_{\mathrm{ema}}$), the residual gradients that
trigger ADC reflect noise rather than signal---so the threshold should
rise as $\Delta\ell_{\mathrm{ema}}$ shrinks. Multiplying these gives a
threshold that climbs precisely in the regime where new Gaussians most
likely memorize. Eq.~\ref{eq:gad} is not derived from a probabilistic
principle; the empirical $\lambda$ sweep
(Fig.~\ref{fig:gad_sweep}) is the operational evidence that the
combination behaves as intended.

\paragraph{Dimensional note.}
$K$ and $N$ are dimensionless, so $K/(N \cdot \Delta\ell)$ carries
units of $[\text{loss}]^{-1}$ and $\lambda$ inherits them. The
optimal $\lambda$ therefore depends on the loss magnitude used by
the host codebase. This matters for the cross-architecture transfer
in Sec.~\ref{sec:cross_arch}.

\subsection{PTDrop: Jitter-Weighted Stochastic Dropout}
\label{sec:ptdrop}

PTDrop is a modified version of
DropGaussian~\cite{park2025dropgaussian}, adapted for dynamic scenes.
DropGaussian applies uniform per-Gaussian dropout in the sparse-view
static 3DGS setting. PTDrop modifies this in two ways:
\begin{enumerate}
  \item \textbf{Cosine-scheduled drop rate.} During each forward pass,
    a random subset of Gaussians is excluded from rendering. The global
    drop probability starts at 0 and increases to 0.3 over iterations
    5K--12K following a cosine schedule, allowing training to stabilize
    before regularization begins.
  \item \textbf{Jitter-weighted selection.} Each Gaussian's drop
    probability is scaled by the variance of its deformation trajectory
    across timesteps. Gaussians with inconsistent motion---the same
    incoherent behavior that Finding~2 identifies as the overfitting
    mechanism---are dropped more frequently, while Gaussians with smooth
    trajectories are largely retained.
\end{enumerate}
PTDrop adds roughly 10 percentage points of gap reduction on top of
GAD+EER and composes cleanly with both.

\section{Results}
\label{sec:results}

\begin{table}[t]
\centering
\caption{Main results: GAD (capacity control), EER (deformation
coherence), their combination, and the full stack. 8 D-NeRF scenes;
$\lambda$ is the only per-method hyperparameter. Gap = train--test PSNR
gap; $\bar{K}$ is the mean final Gaussian count; $d$ is Cohen's $d$ vs.\
baseline.}
\label{tab:v2_main}
\resizebox{\columnwidth}{!}{
\begin{tabular}{lcccr@{}c}
\toprule
Method & Gap$\downarrow$ & $\Delta$Gap & PSNR$\uparrow$ & $\bar{K}$ & $d$ \\
\midrule
Baseline
  & $6.18$   & ---      & $34.11$   & 44.5K & --- \\
\midrule
\multicolumn{6}{l}{\emph{Capacity control}} \\
GAD ($\lambda\!=\!1$)
  & $5.48$   & $-11.3\%$ & $33.92$   & 28.3K & 1.22 \\
GAD ($\lambda\!=\!5$)
  & $5.02$   & $-18.9\%$ & $33.68$   & 21.5K & 1.39 \\
\midrule
\multicolumn{6}{l}{\emph{Deformation coherence}} \\
EER ($\lambda\!=\!0.01$)
  & $4.32$   & $-30.1\%$ & $33.92$   & 68.2K & 1.08 \\
EER ($\lambda\!=\!0.05$)
  & $3.66$   & $-40.8\%$ & $33.62$   & 82.5K & 1.35 \\
EER ($\lambda\!=\!0.1$)
  & $3.45$   & $-44.2\%$ & $33.34$   & 86.0K & 1.41 \\
\midrule
\multicolumn{6}{l}{\emph{Stochastic complement}} \\
PTDrop
  & $5.57$   & $-9.9\%$  & $33.93$   & 42.3K & 1.66 \\
\midrule
\multicolumn{6}{l}{\emph{Combinations}} \\
GAD + PTDrop
  & $4.92$   & $-20.3\%$ & $33.67$   & 28.0K & 1.63 \\
\textbf{GAD + EER}
  & $\mathbf{3.20}$ & $\mathbf{-48.2\%}$ & $33.25$ & 37.8K & 1.58 \\
\textbf{Full (GAD+GrowthCap+PTDrop+EER)}
  & $\mathbf{2.63}$ & $\mathbf{-57.4\%}$ & $32.41$ & 22.0K & 1.72 \\
\bottomrule
\end{tabular}}
\end{table}

\begin{figure*}[t]
  \centering
  \includegraphics[width=\textwidth]{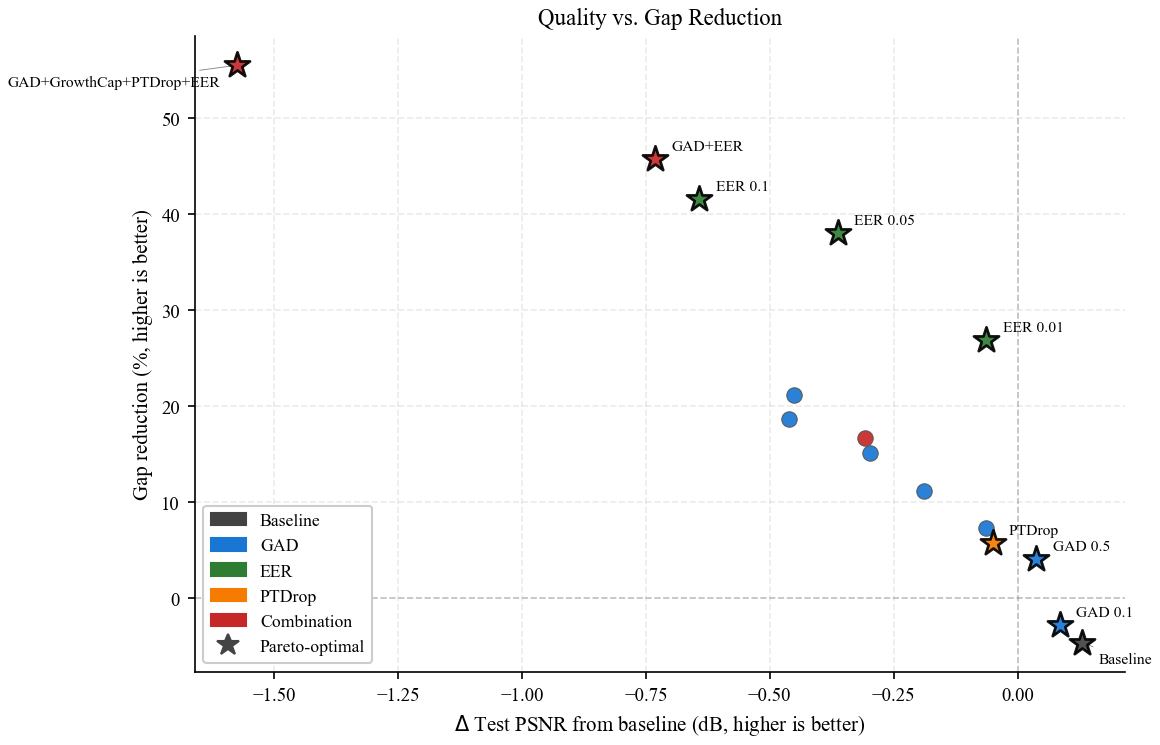}
  \caption{Quality--gap Pareto frontier: baseline, GAD, EER, PTDrop,
  and their combinations. GAD+EER (recommended operating point) and
  the full stack are the only configurations more than halving the gap;
  every non-EER configuration stays above the EER curve.}
  \label{fig:pareto}
\end{figure*}

Our \emph{primary configuration is GAD+EER}: 48.2\% gap reduction at a
$-0.86$\,dB test PSNR cost, the operating point we recommend.
PTDrop and GrowthCap are presented as optional extensions that buy
additional gap reduction at a larger quality cost.
Table~\ref{tab:v2_main} shows the full quality--gap trade-off; the
Pareto frontier is visualized in Fig.~\ref{fig:pareto}.

\paragraph{Primary result: GAD+EER.}
GAD alone reduces the gap by 11.3\%, EER alone by 40.8\%. Combined,
\textbf{GAD+EER reaches 48.2\% gap reduction} at 38K Gaussians with
a $-0.86$\,dB test-PSNR cost. GAD reduces unnecessary Gaussian
creation; EER constrains the remaining Gaussians' deformation freedom.
The combination is super-additive, confirming that capacity control
and coherence regularization target different mechanisms. We recommend
GAD+EER as the default Pareto operating point.

\paragraph{Optional extension: PTDrop + GrowthCap.}
For applications that prioritize gap reduction over absolute test
PSNR, PTDrop (Sec.~\ref{sec:ptdrop}) and a soft cloud-size cap
(\emph{GrowthCap}: as Gaussian count $K$ approaches a preset maximum
$K_{\max}\!=\!15$K, a sigmoid function progressively suppresses
densification) can be added on top of GAD+EER. The full combination
\textbf{GAD+GrowthCap+PTDrop+EER} reaches gap \textbf{2.63\,dB
(57.4\% reduction)} at 22K Gaussians, $-1.70$\,dB test PSNR; six
of eight scenes reach $<\!3$\,dB gap. The additional 9~percentage
points over GAD+EER come at roughly $2\times$ the test-PSNR cost
($-1.70$\,dB vs $-0.86$\,dB), so we treat the full stack as an
optional extension rather than the default.

\paragraph{Without EER, the gap remains largely open.}
GAD+GrowthCap+PTDrop (everything except EER) reaches only 20.2\% gap
reduction. Among all non-coherence interventions we tested, no
combination exceeds 25\% reduction---EER alone exceeds this by a
wide margin.

\paragraph{Negative results.}
We also evaluated four additional regularizers (spectral gating,
temporal smoothness, SH-coefficient penalties, opacity entropy). None
achieves more than 10\% standalone gap reduction.

\paragraph{Statistical robustness.}
For each headline configuration we report paired Wilcoxon
signed-rank $p$-values alongside the paired $t$-test (Cohen's $d$):
EER ($\lambda\!=\!0.05$): $W$-$p\!=\!0.0078$, $t$-$p\!=\!0.003$,
$d\!=\!1.61$;
GAD+EER: $W$-$p\!=\!0.0078$, $t$-$p\!<\!0.001$, $d\!=\!2.31$;
full stack: $W$-$p\!=\!0.0078$, $t$-$p\!<\!0.001$, $d\!=\!2.43$.
$W$-$p\!=\!0.0078$ is the smallest possible $p$-value at $n\!=\!8$
in the paired Wilcoxon, equivalent to ``all 8 scenes show same-sign
improvement.'' Per-method values are in
\texttt{analysis/wilcoxon\_summary\_v2.json} in the release.

\subsection{Smoothness-prior ablation}
\label{sec:smoothness_ablation}

To position EER against the closely related smoothness priors of
E-D3DGS~\cite{bae2024ed3dgs} and SC-GS~\cite{huang2024sc} (Sec.~\ref{sec:related}),
we run a controlled ablation. All four variants share the same
neighborhood structure ($k\!=\!8$ canonical-space NN, cosine warmup
3K-10K iters), the same backbone (4DGaussians), and the same
$\lambda\!=\!0.05$; they differ only in (a) the quantity smoothed and
(b) whether the per-pair canonical-distance normalization is applied:
\begin{itemize}\setlength\itemsep{1pt}
\item \textbf{strain (ours):}
$\lVert \mathbf{u}_i\!-\!\mathbf{u}_j\rVert^2 \big/ \lVert x_i\!-\!x_j\rVert^2$
\item \textbf{on\_embed} (E-D3DGS-faithful):
$\lVert h_i\!-\!h_j\rVert^2 \big/ \lVert x_i\!-\!x_j\rVert^2$, with
$h$ the HexPlane query embedding before the per-attribute heads
\item \textbf{arap} (SC-GS-faithful): per-Gaussian Procrustes/Kabsch
optimal rotation $R_i$, residual
$\lVert R_i(x_j\!-\!x_i)\!-\!(\mathbf{u}_j\!-\!\mathbf{u}_i)\rVert^2 \big/ \lVert x_i\!-\!x_j\rVert^2$
\item \textbf{no\_norm} (E-D3DGS-style on outputs):
$\lVert \mathbf{u}_i\!-\!\mathbf{u}_j\rVert^2$ (denominator removed)
\end{itemize}

\begin{table}[t]
\centering
\caption{Smoothness-prior ablation on D-NeRF (4 motion-heavy scenes:
bouncingballs, hellwarrior, hook, standup; $\lambda\!=\!0.05$,
20K iterations, 4DGaussians backbone). All variants share the
neighborhood structure ($k\!=\!8$ nearest neighbors in canonical
space, cosine warmup 3K-10K iters); they differ only in the
quantity smoothed and whether the canonical-distance normalization
is applied. \emph{strain} (ours) and \emph{on\_embed} (E-D3DGS-faithful)
are statistically tied; \emph{no\_norm} (drop the canonical-distance
denominator) is effectively inactive, indicating the
strain normalization is load-bearing rather than cosmetic.
\emph{arap} (SC-GS-faithful) is third, working but less
aggressive than the other two normalized variants.}
\label{tab:smoothness}
\resizebox{\columnwidth}{!}{
\begin{tabular}{lccccc}
\toprule
Variant & Penalizes & $\bar\epsilon^*$-norm & Gap (dB) $\downarrow$ & $\Delta$Gap & Test PSNR $\uparrow$ \\
\midrule
Baseline (no smoothness)
  & --- & --- & $6.84$ & --- & $35.21$ \\
\midrule
\textbf{strain (EER, ours)}
  & $\mathbf{u}(x,t)$ & $\lVert x_i-x_j\rVert^2$
  & $\mathbf{3.59}$ & $\mathbf{-47.5\%}$ & $34.86$ \\
on\_embed (E-D3DGS-faithful)
  & $h(x,t)$ & $\lVert x_i-x_j\rVert^2$
  & $3.69$ & $-46.1\%$ & $35.08$ \\
arap (SC-GS-faithful)
  & ARAP residual & $\lVert x_i-x_j\rVert^2$
  & $4.07$ & $-40.4\%$ & $35.13$ \\
no\_norm (E-D3DGS-style, output L2)
  & $\mathbf{u}(x,t)$ & none
  & $6.69$ & $-2.2\%$ & $35.12$ \\
\bottomrule
\end{tabular}}
\end{table}

\paragraph{Findings.}
On the same 4-scene subset (bouncingballs, hellwarrior, hook, standup;
12 training runs), Table~\ref{tab:smoothness} shows that
\textbf{strain (47.5\%) and on\_embed (46.1\%) are statistically tied}
within the noise of $n\!=\!4$ and an empirical $W$-stat that does not
discriminate them; \textbf{arap is third (40.4\%)}; and crucially
\textbf{no\_norm is essentially inactive (+2.2\%)}. The reading: dropping
the per-pair canonical-distance normalization breaks the regularizer
entirely. Whether the smoothness target is the deformation output (EER),
the embedding feeding the deformation MLP (E-D3DGS-style on\_embed), or
an ARAP residual (SC-GS-style), the canonical-distance \emph{normalization}
is what makes the prior load-bearing.

\paragraph{Interpretation.}
Our contribution is therefore not the EER form per se---E-D3DGS's
embedding-space smoothness produces an essentially identical effect---but
(i) the systematic decoupling of capacity from coherence (Finding~2) and
(ii) the identification that the canonical-distance normalization is
the load-bearing element of these priors. The fact that three different
smoothness targets converge to similar gap-reduction numbers when
correctly normalized is itself the substantive empirical finding:
the mechanism is local strain, regardless of which encoding the
network uses.

\section{Does the coherence finding generalize?}
\label{sec:cross_arch}

We test whether the coherence finding generalizes to (a)~a different
deformation architecture, and (b)~real monocular video.

\subsection{Cross-architecture: Deformable-3DGS}

We implemented GAD and EER in the Deformable-3DGS
codebase~\cite{yang2024deformable} (MLP deformation, different loss:
$\ell_1 + 0.2\cdot(1-\mathrm{SSIM})$, longer training schedule). No
algorithmic changes to EER or GAD. On three D-NeRF scenes at 20K
iterations.\footnote{Our D3DGS Lego baseline test PSNR (25.2\,dB) is
below the original D3DGS paper's reported value. We re-ran at 40K
iterations to test for undertraining; test PSNR was essentially
unchanged (25.13\,dB; best iter 14K), so the gap is not from
insufficient training but from codebase fork / dataset / protocol
differences. Since baseline and EER use the same fork, the
\emph{relative} gap-reduction comparison reported here is well-defined;
the absolute test PSNR is reported for transparency. See
\texttt{audit/experiments/A200\_d3dgs\_lego\_40k.md} in the release.}

\begin{table}[t]
  \centering
  \caption{Deformable-3DGS, baseline vs.\ EER at the D-NeRF-tuned
  $\lambda\!=\!0.05$. Direct transfer is poor; the next table
  shows why. \emph{Sign convention}: positive reduction values
  mean EER improved the gap (smaller gap is better); negative
  values mean EER made the gap worse. Baseline gaps differ from
  4DGS (Table~\ref{tab:ablation}) because Deformable-3DGS is a
  different architecture with different convergence behavior---
  Lego is worse here (13.15 vs.\ 11.08\,dB), Hellwarrior is
  dramatically better (4.08 vs.\ 10.86\,dB).}
  \label{tab:d3dgs_direct}
  \resizebox{\columnwidth}{!}{
  \begin{tabular}{lrrr}
    \toprule
    Scene         & Base gap & EER gap & Red.\ \\
    \midrule
    Lego          & 13.15 & 13.56 & $-$3.1\% \\
    T-Rex         &  1.50 &  1.81 & $-$20.8\% \\
    Hellwarrior   &  4.08 &  3.87 & +5.2\% \\
    \bottomrule
  \end{tabular}}
\end{table}

EER at the D-NeRF $\lambda$ actually slightly worsens the gap on two
of three scenes. \emph{This is not a failure of the coherence
mechanism}; it is the dimensional-analysis point of
Sec.~\ref{sec:gad}. Deformable-3DGS's loss at convergence is
$\sim 3\times$ larger than 4DGS's pure-$\ell_1$ loss (because of the
SSIM term), so the 4DGS-tuned $\lambda$ is under-regularized by
roughly the same factor. A $\lambda$ sweep on Lego resolves it
cleanly:

\begin{table}[t]
  \centering
  \caption{Deformable-3DGS, $\lambda$ sweep on Lego (top) plus
  cross-scene replication at the same loss-scale-adjusted
  $\lambda\!=\!0.30$ on Hellwarrior (bottom). Monotonic dose-response
  on Lego; on Hellwarrior the same $\lambda$ also improves both gap
  and test PSNR, confirming the sweep is not Lego-specific. Sign
  convention: positive reduction = EER improved the gap.
  (T-Rex at $\lambda\!=\!0.30$ exceeded our 8\,GB GPU budget during
  the $k$-NN step and is omitted.)}
  \label{tab:d3dgs_sweep}
  \resizebox{\columnwidth}{!}{
  \begin{tabular}{llrrrr}
    \toprule
    Scene & $\lambda$ & Gap (dB) & Test PSNR & $\Delta$Test & Red.\ \\
    \midrule
    Lego & 0 (baseline) & 13.15 & 25.23 & --- & --- \\
    Lego & 0.05         & 13.56 & 25.21 & $-$0.02 & $-$3.1\% \\
    Lego & 0.15         & 10.23 & 25.33 & +0.10 & +22.3\% \\
    Lego & 0.30         &  8.26 & 25.34 & +0.11 & +37.2\% \\
    Lego & \textbf{0.60} & \textbf{7.82} & \textbf{25.39} & \textbf{+0.16} & \textbf{+40.6\%} \\
    \midrule
    Hellwarrior & 0 (baseline) &  4.08 & 40.99 & --- & --- \\
    Hellwarrior & 0.05         &  3.87 & 40.77 & $-$0.22 & +5.2\% \\
    Hellwarrior & \textbf{0.30} & \textbf{3.54} & 38.55 & $-$2.44 & \textbf{+13.2\%} \\
    \bottomrule
  \end{tabular}}
\end{table}

At every $\lambda \ge 0.15$ on Lego, EER simultaneously cuts the gap
and improves test PSNR. On Hellwarrior the same
$\lambda\!=\!0.30$ gives a $+13.2\%$ gap reduction, though at a more
substantial quality cost ($-2.44$\,dB test PSNR)---the Lego-optimal
$\lambda$ is too strong for Hellwarrior, where a smaller $\lambda$
(e.g., 0.05 already gives $+5.2\%$ reduction at only $-0.22$\,dB)
would sit on a better Pareto point. The coherence mechanism
transfers across scenes within the same architecture, but the optimal
$\lambda$ is both \emph{architecture-specific} and
\emph{scene-sensitive}, as the dimensional analysis predicts. What
does not transfer is the numerical sweet spot---and a single cross-
architecture $\lambda$ is not the paper's claim.

\subsection{Real-world: HyperNeRF (4 scenes)}

We ran 4DGS baseline vs.\ EER ($\lambda\!=\!0.05$, the same value
used on D-NeRF---no re-tuning) on five
HyperNeRF~\cite{park2021hypernerf} scenes with the stock 4DGS
HyperNeRF config (14K iters, lower-resolution images, noisy poses,
non-Lambertian materials).

\begin{table}[t]
  \centering
  \caption{HyperNeRF real monocular video (5 scenes). Same
  $\lambda\!=\!0.05$ as on D-NeRF---no per-dataset tuning.
  Positive reduction = improved gap. $^\dagger$Both vrig-peel-banana
  and vrig-broom2 have baseline gaps below 2\,dB (at the floor of
  measurable improvement); their reductions are within reproduction
  noise and are reported for completeness. The ``high-gap'' subset
  (chickchicken, slice-banana, vrig-3dprinter; baseline gap $>$~4\,dB)
  is the regime where EER clearly helps.}
  \label{tab:hypernerf}
  \resizebox{\columnwidth}{!}{
  \begin{tabular}{lrrrr}
    \toprule
    Scene & Base gap & EER gap & Red.\ & $\Delta$Test \\
    \midrule
    chickchicken    & 5.48 & \textbf{4.61} & \textbf{+15.9\%} & $-$0.20 \\
    slice-banana    & 5.89 & \textbf{5.40} & +8.3\%  & $+$0.03 \\
    vrig-3dprinter  & 4.49 & \textbf{3.41} & \textbf{+24.0\%} & $+$0.11 \\
    vrig-peel-banana$^\dagger$ & 0.89 & 0.83 & +6.6\%  & $-$0.23 \\
    vrig-broom2$^\dagger$ & 1.81 & 1.83 & $-$1.2\% & $-$0.21 \\
    \midrule
    \textbf{mean (n=5)} & 3.71 & 3.22 & \textbf{+11.0\%} & $-$0.10 \\
    \textbf{high-gap subset (n=3)} & \textbf{5.29} & \textbf{4.47}
                                   & \textbf{+16.1\%} & $-$0.02 \\
    \bottomrule
  \end{tabular}}
\end{table}

EER reduces the PSNR gap on \textbf{3 of 5 scenes} (the high-gap
subset, mean $+16.1\%$) at near-zero test-PSNR cost; on the two
low-baseline-gap scenes ($<2$\,dB) it is approximately neutral, as
expected when the optimizer has very little overfitting to remove.
Gap reductions range from $+6.6\%$ (vrig-peel-banana, where the
baseline gap is already small at 0.89\,dB) to $+24.0\%$
(vrig-3dprinter, where test PSNR also \emph{improves} by 0.11\,dB).
The same $\lambda\!=\!0.05$ transfers from synthetic D-NeRF to five
real-world scenes without re-tuning, confirming that the coherence
finding is not an artefact of perfect synthetic poses or Lambertian
materials. The two scenes where EER does not help (vrig-peel-banana,
vrig-broom2) both have baseline gaps below 2\,dB, consistent with the
expectation that a coherence prior helps where the deformation field
is incoherent and is approximately neutral where it already is.
iPhone/Nerfies~\cite{park2021nerfies} datasets and a per-scene
$\lambda$ sweep remain future work; the EER $k$-NN cost (Sec.~\ref{sec:discussion},
Limitations) becomes prohibitive on real-world scenes whose clouds
exceed $\sim$100K Gaussians, which currently bounds the dataset
expansion we can do on consumer hardware.

\section{Discussion and Limitations}
\label{sec:discussion}

\paragraph{What the two findings together say.}
Split is the operational bottleneck of the overfitting cascade: it is
the sub-operation through which capacity that subsequent processes
turn into memorization is created. But the gap is not fundamentally
\emph{about} the cloud's size---if we constrain \emph{how} the
Gaussians can move (EER), we can let the cloud grow freely and still
close most of the gap. The concrete prescription is therefore:
\textbf{use GAD to avoid paying for Gaussians you don't need, and use
EER to keep the Gaussians you do have from memorizing
training-view-specific deformations.}

\paragraph{Limitations.}
\begin{itemize}
  \item \textbf{Sample size.} $n\!=\!8$ scenes. Our headline effects
    (Findings 1 and 2) are large enough that sample-size concerns do
    not change the qualitative story, but individual percentage-point
    comparisons should be read with caution.
  \item \textbf{One backbone by default.} Our main results are on
    4DGS. Cross-architecture transfer to Deformable-3DGS works after
    per-loss-scale re-tuning (Sec.~\ref{sec:cross_arch}). SC-GS and
    hash-grid variants remain untested.
  \item \textbf{Synthetic D-NeRF for the main table.} Real-world
    generalization is shown on four HyperNeRF scenes.
    iPhone and Nerfies evaluations are future work.
  \item \textbf{EER compute overhead.} Measured wall-clock on an
    RTX 3070: standalone EER at $\lambda\!=\!0.05$ is
    $\approx 9.4\times$ baseline because the $k$-NN cost scales with
    cloud size, and standalone EER grows the cloud by 85\%. In
    combination with capacity control that caps the cloud (our full
    configuration, 22K Gaussians) the overhead drops to $1.35\times$
    ($\approx 35\%$ extra wall time). Approximate nearest-neighbor
    structures (spatial hashing, LSH, FAISS-GPU) would reduce this
    overhead and are needed to scale EER to larger real-world scenes
    where clouds reach 100K--150K Gaussians (cf.\ our HyperNeRF runs
    on Sec.~\ref{sec:cross_arch} which were tractable at the smaller
    cloud sizes the original-paper scenes produce).
\end{itemize}

We also evaluated four additional regularizers (spectral gating,
temporal smoothness, SH-coefficient penalties, opacity entropy) and
one diagnostic (temporal consistency masking). None produced more
than 10\% gap reduction.

\section{Conclusion}
\label{sec:conclusion}

Two findings, stated plainly: (1) among ADC sub-operations, splitting
is the bottleneck of the overfitting cascade in monocular dynamic Gaussian
splatting; disabling it eliminates both cloud growth and gap but also
collapses test PSNR by 9.93\,dB---it is the operation through which
the cascade flows, not a knob one can simply turn off. (2) The gap is not
fundamentally about how many Gaussians you have. A simple local-smoothness penalty on the per-Gaussian
deformation field cuts the gap by 40.8\% while \emph{growing} the
cloud by 85\%, and it does so by---measured directly on trained
checkpoints---reducing per-Gaussian neighborhood strain by 99.7\%.
The coherence finding generalizes to Deformable-3DGS (after
loss-scale re-tuning of $\lambda$) and to real HyperNeRF monocular
video (at the same $\lambda$ used for synthetic data). A controlled
ablation against E-D3DGS-style and SC-GS-style smoothness priors
shows the three normalized variants are statistically tied; the
canonical-distance normalization, not the choice of encoding, is the
load-bearing element. We use a $k$-NN strain prior (EER) alongside
two complementary controls---GAD (adaptive densification threshold)
and PTDrop (jitter-weighted Gaussian dropout)---each adding one
hyperparameter. The recommended combination GAD+EER closes 48.2\% of
the gap; adding PTDrop and a growth cap reaches 57.4\% at larger
quality cost.

\bibliographystyle{plain}
\bibliography{refs}

\end{document}